\documentclass
{article}
\usepackage{amsmath, amssymb}
\usepackage[numbers]{natbib}
\usepackage{tikz-cd}

\newtheorem{Definition}{Definition}[section]
\newtheorem{theorem}{Theorem}[section]

\begin{document}

\title{\Large An Application of HodgeRank to Online Peer Assessment
}
\author{Tse-Yu Lin\thanks{Department of Mathematical Sciences, National Chengchi University.} \\
\and
Yen-Lung Tsai\thanks{Department of Mathematical Sciences, National Chengchi University.}}
\date{}

\maketitle







\begin{abstract} \small\baselineskip=9pt 
Bias and heterogeneity in peer assessment can lead to the issue of unfair scoring in the educational field. To deal with this problem, we propose a reference ranking method for an online peer assessment system using HodgeRank. Such a scheme provides instructors with an objective scoring reference based on mathematics.
\end{abstract}

\section{ Introduction}
In this paper, we construct a reference score for online peer assessments based on HodgeRank~\cite{jiang2011statistical}. Peer assessment is a process in which students grade their peers’ assignments~\cite{falchikov2000student, topping1998peer}.

A peer assignment system is used to enhance students’ learning process, especially in higher education. Through such a system, students are given the opportunity to not only learn knowledge from textbooks and instructors, but also from the process of making judgements on assignments completed by their peers. This process helps them understand the weaknesses and strengths in the work of others, and then to review their own.

However, there are some practical issues associated with a peer assignment system. For example, students tend to give significantly higher grades than senior graders or professionals (see ~\cite{freeman2010accurate} for more details). Also, students have a tendency to give grades within a range, with the center of such a range often being based on the first grade they gave. Therefore, bias and heterogeneity can occur in a peer assignment system.

There are various ranking methods on peer assessment problem, such as PeerRank~\cite{walsh2014} and Borda-like aggregation algorithm~\cite{caragiannis2015}. PeerRank, a famous method based on a iterative process to solve the fixed-point equation. PeerRank has many interesting properties from the view of linear algebra. Borda-like aggregation algorithm, a random matheod based on the theory of random graphs and voting theory, whcih provides some probabilistic explanation on peer assessment problem.

In this paper, we propose another ranking scheme to deal with peer assessment problems that uses HodgeRank, a statistical preference aggregation problem from pairwise comparison data. The purpose of HodgeRank is to find a global ranking system based on pairwise comparison data. HodgeRank can not only generate a ranking order, but also highlight inconsistencies in the comparisons (see~\cite{jiang2011statistical} for more detail). We apply HodgeRank to the problems in online assessment and display ranking results from HodgeRank and PeerRank in turn.

We will briefly introduce HodegRank and its useful properties in next section.

%

\section{HodgeRank}

HodgeRank, a statistical ranking method based on combinatorial Hodge theory to find a consistent ranking. Rigorously speaking, HodgeRank is one solution of a graph Laplacian problem with minimum Euclidean norm.

Now, we start from notations borrowed from graph theory. 

Consider a connected graph $\mathcal{G} = (V, E)$, where $V=\{1, 2, \cdots, n\}$ is the set of alternatives to be ranked, and $E\subseteq V\times V$, consists of some unordered pairs from $V$. 

In this paper, $V$ represents the set of students to be ranked by their peers, and $E$ collects the information of pairwise comparisons. i.e., $(i, j)\in E$ if students $i$ and $j$ are compared at least once.

Denote $\Lambda$ to be the number of assignments. Then for each assignment $\alpha\in\Lambda$, pairwise comparison data on a graph $\mathcal{G}$ of assignment $\alpha$, is given by $Y^\alpha:E\to\mathbb{R}$ so that $Y^\alpha$ is skew-symmetry. i.e., $Y^\alpha_{ij} = - Y^\alpha_{ji}$ for all $i,j\in V$. $Y^\alpha_{ij}>0$ if grade of the student $j$ is higher than student $i$ by $Y^\alpha_{ij}$ credits. For example, $Y^\alpha_{ij}\in[-100, 100]$ on hundred-mark system.

For each $\alpha\in\Lambda$, a weight matrix $W^\alpha = [w_{ij}^\alpha]$ is associated as follows: $w_{ij}^\alpha>0$ if $Y_{ij}^\alpha\neq0$, and $0$ otherwise. Set $W = \sum\limits_{\alpha\in\Lambda}W^\alpha$.

Let $Y = \sum\limits_{\alpha\in\Lambda}Y^\alpha$ be a $n$-by-$n$ matrix. The goal of the HodgeRank is find a ranking $s:V\to\mathbb{R}$ so that 
\begin{equation} \label{e1.1}
Y_{ij} = s_j - s_i\mbox{ for all }i,j\in V.
\end{equation}

However, equations (\ref{e1.1}) can not be admissible in general. Consider the following example,
\[
Y = \begin{bmatrix}
0 & 1 & -1\\
-1 & 0 & -1\\
1 & 1 & 0
\end{bmatrix}
\]
If there exists $s:V\to\mathbb{R}$ such that (\ref{e1.1}) hold. Then 
\[
1 = Y_{12} = s_2-s_1 = (s_2-s_3)+(s_3-s_1) = Y_{32} + Y_{13} = 0
\] 
which leads to a contradiction. That is, it is impossible to solve (\ref{e1.1}) for any skew-symmetric matrix $Y$. Therefore, we should consider the least square solution of (\ref{e1.1}) instead. Before we rewrite above problem, we need to introduce some notations below. 

\begin{Definition}{\rm ~\cite{jiang2011statistical} Denote
\[\mathcal{M}_G = \{X\in\mathbb{R}^{n\times n}~|~X_{ij} = s_i-s_j\mbox{for some }s:V\to\mathbb{R}\},\] the space of global ranking,
and the combinatorial gradient operator
\[
\mbox{grad}: \mathcal{F}(V, \mathbb{R})\to \mathcal{M}_G
\]
is an operator defined from $\mathcal{F}(V, \mathbb{R})$, the set of all function from $V$ to $\mathbb{R}$ (or the space of all potential functions), to $\mathcal{M}_G$, as follows
\[
\big(\mbox{grad}s\big)(i, j) = s_j - s_i.
\]
}
\end{Definition}

From the example above, it is easy to find that if $X = grad(s)$ for some $s\in\mathcal{F}(V, \mathbb{R})$, then $X_{ij}+X_{jk}+X_{ki} = 0$ for any $(i, j), (j, k), (k, i)\in E$. However, the converse might not be true in general. That is, denote 
\[
\mathcal{A}=\{X\in\mathbb{R}^{n\times n}~|~X^T=-X\},
\] the set of all skew-symmetric matrices, and let 
\[
\mathcal{M}_T=\{X\in\mathcal{A}~|~X_{ij}+X_{jk}+X_{ki}=0\},
\]
then $\mathcal{M}_G\subseteq\mathcal{M}_T$.

With these notations above, then the above problem becomes the following optimization problem:

\[
\min\limits_{X\in\mathcal{M}_G}|| X - Y||^2_{2, w}
=
\min\limits_{X\in\mathcal{M}_G}\sum\limits_{(i, j)\in E}w_{ij}(X_{ij}-Y_{ij})^2
\]

That is, once a graph is given, then the weight on edge $E$ determines an optimization problem. Conversely, a graph can intuitively arise from the ranking data. 

Let $\{Y^\alpha~|~\alpha\in\Lambda\}$ be a set of $n$-by-$n$ skew-symmetric matrices, and $\{W^\alpha~|~\alpha\in\Lambda\}$ is associated as above. 

Then an undirected graph $\mathcal{G}=(V, E)$ can be defined by $V = \{1, 2, \cdots, n\}$ and 
\[
E = \{(i, j)\in V\times V~|~W_{ij}>0\}.
\]
In this case, we can treat $X$ as a edge flow on $\mathcal{G}$ in the sense of combinatorial vector calculus.

In conclusion, we have the following relation between graph and

\[
\begin{tikzcd}
\mathcal{G}=(V, E)\arrow[rr, Leftrightarrow] & & \left\{\begin{tabular}{l}
$X^T = -X$\\
$W = \sum\limits_{\alpha\in\Lambda}W^{\alpha}$.
\end{tabular}\right.
\end{tikzcd}
\]

Hence, the optimization problem of a skew-symmetric least square problem can be view as an optimization problem of edge flow on a graph.

\begin{Definition}(Consistency){\rm ~\cite{jiang2011statistical} 
Let $X:V\times X\to\mathbb{R}$ be a pairwise ranking edge flow on a graph $\mathcal{G}=(G, E)$.
\begin{itemize}
\item X is called consistency on $\{i, j, k\}$ if 

$(i,j), (j,k), (k,i)\in E$ and $X\in\mathcal{M}_T$ 
\item X is called globally consistency on $\{i, j, k\}$ if $X=\mbox{grad}(s)$ for some $s\in\mathcal{F}(V,\mathbb{R})$
\end{itemize}
}
\end{Definition}

Note that if $X$ is called globally consistency, then $X$ is consistency on any 3-clique $\{i, j, k\}$, where $(i,j), (j,k), (k,i)\in E$.

Now, consider the weighted trace induced by $W$. i.e., 

\[
<X, Y>=\mbox{tr}\big(X^T(W\odot Y)\big)=\sum\limits_{(i,j)\in E}W_{ij}X_{ij}Y_{ij}
\] for $X,Y\in\mathcal{A}$, where $\odot$ represents the Hadamard product or elementwise product.

With this weighted inner product, we obtain two orthogonal complement of $\mathcal{A}$
\[
\mathcal{A} = \mathcal{M}_G\oplus \mathcal{M}_G^{\perp}
= \mathcal{M}_T\oplus \mathcal{M}_T^{\perp}
\]

Since $\mathcal{M}_G\subseteq\mathcal{M}_T$, we have $\mathcal{M}_G^{\perp}\supseteq\mathcal{M}_T^{\perp}$ and we can get further orthogonal direct sum decomposition of $\mathcal{A}$ as follows:
\[
\mathcal{A} = \mathcal{M}_G\oplus \mathcal{M}_H\oplus \mathcal{M}_T^{\perp},
\]
where $\mathcal{M}_H=\mathcal{M}_T\cap\mathcal{M}_G^{\perp}$.

This decomposition is called the combinatorial Hodge decomposition. For more detail about the theory of combinatorial Hodge decomposition, please refer~\cite{jiang2011statistical} for more detail.

We now state one useful theorem in~\cite{jiang2011statistical}.

\begin{theorem}{\rm ~\cite{jiang2011statistical}\label{t2.1}
\begin{enumerate}
\item The minimum norm solution $s$ of (\ref{e1.1}) is the solution of the normal equation:
\[
\Delta_0 s = -\mbox{div}~Y,
\]
where $\Delta_0=\left\{\begin{tabular}{ll}
$\sum\limits_{(i,j)}w_{ij}$ & if $i = j$\\
$-w_{ij}$ & if $j\in V$ with $(i,j)\in E$\\
0 & otherwise
\end{tabular}\right.$, and

\[
\mbox{div}(Y)(i)=\sum\limits_{j s.t. (i,j)\in E}w_{ij}Y_{ij}
\]
is the combinatorial curl operator of $Y$.

\item The minimum norm solution $s$ of (\ref{e1.1}) is 
\[
s^*=-\Delta_0^\dagger~\mbox{div}Y,
\]
where $\Delta_0^\dagger$ represents the Moore-Penrose pseudo inverse  of the matrix $\Delta_0$.
\end{enumerate}
}
\end{theorem}

The Hodge decomposition indicates the solution of $(\ref{e1.1})$, while the theorem~\ref{t2.1} shows how to calculate the minimum solution by solving the normal equation. In the next section, we display how to apply HodgeRank to the online peer assessment problem.


\section{Online peer assessment problem}

As previously mentioned, bias and heterogeneity can lead to unfair scoring in online peer assessments. Students usually grade other students based on the first score they gave, which causes bias. However, since scores are usually compared with others, we can use this comparison behavior to reconstruct true ranking.

The data we used in this section were collected from an undergraduate calculus course. In this course, 133 students were asked to upload their GeoGebra ~\cite{hohenwarter2002geogebra} assignments. Each student was then asked to review five randomly chosen assignments completed by their peers to receive partial credits in return. There are 13 assignments during one semester.

Note that ne key point of the HodgeRank is the connectedness of the graph generated by pairwise comparison data. From table ~\ref{table1.1} above, we can easily see that after half the semester passed, comparison data between students forms a connected graph. Hence, we can apply HodgeRank to calculate the ranking of all the students after assignment 7.

\begin{table}[h]
\caption{Number of components with respect to the number of assignments}
\centering
\begin{tabular}{|c|ccccccc|}\hline
Assignment \# & 1 & 2 & 3 & 4 & 5 & 6 & $7\sim13$\\\hline
\# of components & 21 & 5 & 4 & 3 & 2 & 2 & 1\\\hline
\end{tabular}\label{table1.1}
\end{table}

The traditional method for finalizing peer assessment consists of either using an average cumulative score or a truncated average score. Although these approaches might have some statistical meaning, they cannot avoid bias and heterogeneity in peer assessment.

\begin{figure}[h]
\centering
\includegraphics[width=0.6\textwidth]{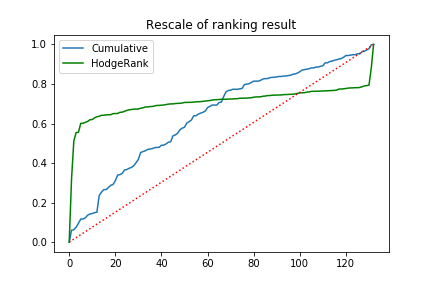}
\caption{Final results using different ranking methods}\label{fig1.1}
\end{figure}

Figure~\ref{fig1.1} displays the cumulative score, PeerRank and HodgeRank, respectively. Here, $(\alpha, \beta) = (0.5, 0)$ in the setting of PeerRank. For what these parameters represent in PeerRank, please refer to ~\cite{walsh2014} for more discussion.

To compare these results, ranking results were normalized into the interval [0, 1] linearly and sorted in ascending order. In addition, to reveal the tendency of each ranking method, a steady line was plotted on the graph. There are some interesting implications that can be observed from this figure. 

First, the cumulative score offers a ranking higher than the steady line. This reflects the existence of bias and heterogeneity in the cumulative average method. Second, PeerRank can be viewed as a modification of the average scoring. Third, sorted ranking result from HodgeRank is a normal distributed curve. This result can might be an explanation why HodgeRank can be solution to eliminate bias and heterogeneity by the normality.

Note that the reason why HodgeRank and PeerRank show different results is their conclusion base are totally different, while former method relies on the pairwise comparison data and latter one is applied on the average score as an initial ranking. Hence, HodgeRank provides instructors with an objective scoring reference using score difference rather than cumulative or average score.

In conclusion, this is the first time HodgeRank has been applied in the field of education. While numerical results were processed using real world data in this study, certain issues, such as how to aggregate the HodgeRank ranking method into a peer assessment system, remain unsolved. This task will be attempted as part of our future work.

\end{document}